# Metalinguistic Information Extraction for Terminology

**Carlos Rodríguez Penagos**
Language Engineering Group, Engineering Institute
UNAM, Ciudad Universitaria A.P. 70-472
Coyoacán 04510  Mexico City, México
CRodriguezP@iingen.unam.mx

**Abstract**

This paper describes and evaluates the Metalinguistic Operation Processor (MOP) system for automatic compilation of metalinguistic information from technical and scientific documents. This system is designed to extract non-standard terminological resources that we have called Metalinguistic Information Databases (or MIDs), in order to help update changing glossaries, knowledge bases and ontologies, as well as to reflect the metastable dynamics of special-domain knowledge.

## 1 Introduction

Mining terminological information from free or semi-structured text in large-scale technical corpora is slowly becoming a reasonably mature NLP technology, with term extraction systems leading the way. Automatically obtaining information *about* terms from free text has been a field less explored, but recent experiences have shown that compiling the extensive resources that modern scientific and technical disciplines need to manage the explosive growth of their knowledge is both feasible and practical. A good example of this NLP-based processing need is the National Library of Medicine's MedLine abstract database, which incorporates around 40,000 new Life Sciences papers each month. In order to maintain and update UMLS knowledge resources[1] the NLM staff needs to manually review 400,000 highly-technical papers each year (Powell et al. 2002). Most of these terminological knowledge sources have been compiled from existing glossaries and vocabularies that might become dated fairly quickly, and elucidating this information from domain experts is not an option. Neology detection, terminological information update and other tasks can benefit from automatic search, in highly technical text, of semantic and pragmatic information, e.g. when new information about sublanguage usage is being put forward. In this paper we describe and evaluate the Metalinguistic Operation Processor (MOP) system, implemented to automatically create Metalinguistic Information Databases (or MIDs) from large collections of special-domain research and reference documents. Section 2 discusses previous work, while Section 3 provides an overview of metalinguistic exchanges between experts, and their role in the constitution of technical knowledge. Section 4 presents experiments to localize and disambiguate good candidate metalinguistic sentences, using rule-based and stochastic learning strategies. Section 5 focuses on the problem of identifying and structuring the different linguistic constituents and surface segments of metalinguistic predications. Finally, Section 6 offers a discussion of results and suggestions for possible applications and future lines of research.

## 2 Previous work

One of the constraints of recent lines of research (Pearson, 1998; Klavans et al., 2001; Pascual & Pery-Woodley, 1997) is their focus on definitions, a theoretical object that, although undoubtedly useful and extensively described, presents by its very nature certain limitations when studying expert-domain peer-to-peer communication.[2] The meaning normalization process inherent in compiling definitions may be desirable when creating human-readable reference sources, but might lead to a loss of valuable information for specific contexts where the term appears. Pragmatic information (valid usage conditions or contextual restriction for the terms), or purely evaluative statements (usefulness or validity of a certain term for its intended purpose), might not be found in *classical* definitional contexts. Metalinguistic information in texts can provide us with information not only about what terms mean, but also how they are actually used by domain experts. A wide spectrum of sentential realizations of these kinds of information has been reported by Meyer (2001) and Rodríguez (2001), and organizing it to provide useful terminological resources is left for manual review by human lexicographers. We believe that using the more general concept of metalanguage can automate as much as possible the extraction of fine-grained knowledge about terms, as well as better capture the dynamical nature of the evolution of the scientific and technical knowledge created through the interaction of expert-domain groups.

---

[1] The MeSH and SPECIALIST vocabularies, a Metathesaurus, a Semantic Network, etc.

[2] In some recent approaches, Meyer (2001) and Condamines & Rebeyrolles (2001) exploit wider lexico-conceptual relations in free-text that can be difficult to model and locate accurately.

## 3 Metalanguage, terminology and scientific knowledge

### 3.1 Corpora used in our research

Preliminary empirical work to explore how researchers modify the terminological framework of their highly complex conceptual systems included an initial manual review of 19 sociology articles (138k words) in academic journals. We looked at how term introduction and modification was done, as well as how metalinguistic activity was signalled in text, both by lexical and paralinguistic means. Some of the indicators found included verbs and verbal phrases like *called, known as, defined as, termed, coined, dubbed*, and descriptors such as *term* and *word*. Non-lexical markers included quotation marks, apposition and text layout.[3] The metalinguistic patterns thus identified were expanded (using variations of lexemes, verbal tenses and forms) into 116 queries to the scientific and learned domains of the British National Corpus. The resulting 10,937 sentences (henceforth, the MOP corpus) were manually classified as metalinguistic or otherwise, with 5,407 (49.6% of total) found to be truly metalinguistic sentences, using the criteria described in Section 3.2 below.[4] Other corpora from different domains (described in Section 4) was used both in this preliminary analysis of metalinguistic exchanges, as well as in evaluation and development of the MOP system.

### 3.2 Explicit Metalinguistic Operations

Careful analysis of these corpora, as well of examples in other European languages, presented some interesting facts about what we have termed "Explicit Metalinguistic Operations" (or EMOs):[5]

A) EMOs do not usually follow the *genus-differentia* scheme of aristotelian definitions, nor conform to the rigid and artificial structure of lexicographic entries. More often than not, specific information about language use and term definition is provided by sentences such as (1), in which the term *trachea* is linked to the description *fine hollow tubes* in the context of a globally non-metalinguistic sentence:

(1) This means that they ingest oxygen from the air via fine hollow tubes, known as tracheae.

In research papers partial and heterogeneous information is much more common than complete definitions, although it might otherwise in textbooks geared towards learning a discipline.

B) Introduction of metalinguistic information in discourse is highly regular, regardless of the domain. This can be credited to the fact that the writer needs to mark these sentences for special processing by the reader, as they dissect across two different semiotic levels: a meta-language and its *object* language, to use the terminology of logic where these concepts originated.[6] Their constitutive *markedness* means that most of the times these sentences will have at least two indicators of metalinguistic nature. These formal and cognitive properties of EMOs facilitate the task of locating them accurately in text.

---

[3] Similar work by Pearson (1998) obtained many of the same patterns from the Nature corpus of exact science documents.

[4] Reliability of human subjects for this task has not been reported in the literature, and was not evaluated in our experiments.

[5] We have used the term to highlight the operational nature of such textual instances in technical discourse.

[6] Natural language has to be split (at least methodologically) into two distinct systems that share the same rules and elements: a metalanguage used to refer to an object language, which in turn can refer to and describe objects in the mind or in the physical world. The fact that the two are isomorphic accounts for reflexivity, the property of referring to itself, as when linguistic items are mentioned instead of being used normally in an utterance. Rey-Debove (1978)

C) EMOs can be further analyzed into 3 distinct components, each with its own properties and linguistic realizations:

i) An **autonym** *(see note 6)*: One or more self-referential lexical items that are the logical or grammatical subject of a predication.

ii) An **informational segment**: a contribution of relevant information about the meaning, status, coding or interpretation of a linguistic unit. Informational segments constitute what we state about the autonymical element.

iii) **Markers/Operators:** Elements used to make prominent the whole discourse operation and its non-referential, metalinguistic nature. They are usually lexical, paralinguistic or pragmatic devices that articulate autonyms and informational segments into a predication.

In a sentence such as (2) we have marked the autonym with *italics*, the informational segment with **bold** type and the marker-operator items with square brackets:

(2) **The bit sequences representing quanta of knowledge** [ will be called " ] *Kenes* [ " ], **a neologism intentionally similar to 'genes'** .

### 3.3 Knowledge and knowledge of language

Whenever scientists advance the state of the art of a discipline, their language has to evolve and change, and this build-up is carried out under metalinguistic control. Previous knowledge is transformed into new scientific common ground and ontological commitments are introduced when semantic reference is established. That is why when we want to structure and acquire new knowledge we have to go through a resource-costly cognitive process that integrates within coherent conceptual structures and theories a considerable amount of new and very complex lexical items and terms. Technical terms are not, by definition, part of the far larger linguistic competence of a first native language. Unlike everyday words within a specific social group,

---

follows Carnap in calling this condition *autonymy*.

terms are conventional, even if they have derived from a word that originally belonged to collective competence. We could even posit that all technical terms owe their existence to a baptismal speech act, and that given a big enough sample (an impossibly exhaustive corpus of all expert language exchanges), an initial metalinguistic sentence could be located that constitutes an original, foundational source of meaning.

The information provided by metalinguistic exchanges is not usually inferable from previous one available to the speaker's community, and does not depend on general language competence by itself, but nevertheless is judged important and relevant enough to warrant the additional processing effort involved. Computing what is relevant metalinguistic information has to be done dynamically by figuring out which terminological items can be assumed to be shared by all, and which are new or have to be modified. It's an extended and more complex instance of lexical alignment between interlocutors (Pickering & Garrod, in press). Observing closely how this alignment is achieved can allow us to create computer applications that mimic some aspects of our impressive human competence as efficient readers of technical subjects, as incredibly good lexical-data processors that constantly update and construct our own special purpose vocabularies.

### 4 Filtering out non-metalinguistic sentences: two NLP approaches

The first issue to tackle when mining metalanguage is how to obtain a reliable set of candidate sentences for input into the next extraction phases. We employ a "discourse-oriented" approach that differs from Meyer's (2001) "term-oriented" one. We do not assume we have initially identified a terminological unit and proceed from there, but rather we first locate a metalinguistic discourse operation where a term can be retrieved along with information that refers to it. Condamines & Rebeyrolles (2001) and Meyer (2001) both exploit patterns of "knowledge-rich contexts" to obtain semantic and conceptual information about terms, either to inform terminological definitions or provide structure for a terminological system. A key problem in such approaches that use lexical-based "triggers" is how to control the amount of "noise", or non-relevant instances. The experiments in this section compare two different NLP techniques for this task: symbolic and statistic techniques.

From our initial analysis of various corpora we selected 44 patterns that showed the best statistical reliability as EMO indicators.[7] We started out by tokenizing text, which then was run through a cascade of finite-state devices that extracted a set of candidate sentences before filtering out non-metalinguistic instances. Our filtering distinguishes between useful results, e.g. using the lexical pattern *called* in (3) from non-metalinguistic instances in (4):

(3) Since the shame that was elicited by the coding procedure was seldom explicitly mentioned by the patient or the therapist, Lewis **called** it unacknowledged shame.

(4) It was Lewis (1971;1976) who **called** attention to emotional elements in what until then had been construed as a perceptual phenomenon .

We experimented with two strategies for disambiguation: first, we used collocations as added restrictions (e.g., verbal vs. nominal occurrences of our lexical markers) to discard non-metalinguistic instances, for example *attention* in sentence (4) next to the marker *called*. The next table shows a sample of the filtering collocations.

| Preceding | Subsequent |
|---|---|
| for ***calls*** | |
| in, duty, personal, conference, local, next, the, their, house, anonymous, phone, telephone... | out, someone, charges, before, charge, back, contact, for, upon, to, into, off, 911, by... |
| for ***coin*** | |
| pound, small, pence, in, toss, the, this, a, that, one, gold, silver, metal, esophageal ... | toss |

We also implemented learning algorithms trained on a subset from our EMO corpus, using as vectors either Part-of Speech tags or word strings, at one, two, and three positions adjacent before and after our lexical markers. Our evaluations are based on 3 document sets: a) our original exploratory sociology corpus [5,581 sentences, 243 EMOs]; b) an online histology textbook [5,146 sentences, 69 EMOs]; and c) a small sample from the MedLine abstract database [1,403 sentences, 10 EMOs]. Our system is coded in Python, using the NLTK platform (nltk.sf.net) and a Brill tagger by Hugo Liu at MIT.

### 4.1 The collocation-based approach

Our first approach fared well, with good precision numbers but not so encouraging recall. The sociology corpus gave 0.94 Precision (P) and 0.68 Recall (R), while the histology one presented 0.9 P and 0.5 R. These low recall numbers reflect the fact that we used a non-exhaustive list of metalinguistic patterns. Example (5) shows one kind of metalinguistic sentence attested in corpora that the system does not extract or process:

(5) "Intercursive" power, on the other hand, is power in Weber's sense of constraint by an actor or group of actors over others.

We also tested extraction against a golden standard where sentences that had patterns that our list was not designed to retrieve were removed, which gave a more realistic picture of how the extraction system worked for the actual dataset it was designed to consider. For the sociology corpus (and a ß factor of 1), P was 0.97 and R 0.79, with an F-measure of 0.87. In the histology one P was measured at 0.94, R at 0.81 and F-measure at 0.87. In order to better compare the two filtering strategies, we decided also to zoom in on a more limited subset of verb forms (namely, *calls, called, call*), which presented ratios of metalinguistic relevance in our MOP corpus ranging from 100% positives (for the pattern *so called* + quotation marks) to 31% (*call*). Restricted to these verbs, our metrics showed precision and recall rates around 0.97. One problem with this approach is that the hand-coded rules are domain-specific, and customization for other domains is labour-intensive. In our tests, although most of the collocations work language-wide (phrasal verbs or prepositions), some of them are very specific.[8] Although collocation-based filtering will result in a working system, customization is error-prone and laborious.

### 4.2 Testing learning algorithms

We selected the co-text of marker/operators as relevant features for classifiers based on well-known naive Bayes and Maximum Entropy algorithms that have been reported to work well

---
[7] We excluded dispositional and typographical clues from our selectional patterns, involving mainly lexica and punctuation.

[8] "esophageal coins" is quite unusual outside of medical documents.

with sparse data.[9] We used either as grammatical context the POS tags or the word forms immediately adjacent in one to three positions before and after our triggering markers. Testing all possible combinations evaluates empirically the ideal mix of algorithm, feature type and coverage that insures best accuracy. The naive Bayes algorithm estimates the conditional probability of a set of features given a label, using the product of the probabilities of the individual features given that label. It assumes that the feature distributions are independent, but it has been shown to work well in cases with high degree of feature dependencies. The Maximum Entropy model establishes a probability distribution favouring entropy or uniformity subject to the constraints encoded in the feature-known label correlation. To train our classifiers, Generalized and Improved Iterative Scaling algorithms were used to estimate the optimal maximum entropy of a feature set, given a corpus.[10] 1,371 training sentences from our MOP dataset were converted into YES-NO labelled vectors. The following example from the textual segment *"... creates what Croft calls a description constraint ..."*, uses 3 positions and POS tags:

('VB WP NNP', 'calls', 'DT NN NN')/'YES'@[102].

The different number of positions to the left and right of our training sentences, as well as the nature of the features selected (there are many more word-types than POS tags) ensured that our 3-part vector introduced a wide range of features against our 2 possible labels. The best results of each algorithms restricted to the lexeme *call*, are presented in the next table. Figures 1 and 2 present best results in the learning experiments for the complete set of patterns used in the collocation approach, over two of our evaluation corpora.[11]

| Type | Positions | Tags/Words | Features | Accuracy | Precision | Recall |
|------|-----------|------------|----------|----------|-----------|--------|
| GIS  | 1 | W | 1254 | 0.97 | 0.96 | 0.98 |
| IIS  | 1 | T | 136  | 0.95 | 0.96 | 0.94 |
| NB   | 1 | T | 136  | 0.88 | 0.97 | 0.84 |

---

[9] *see* Rish, 2001, Ratnaparkhi, 1997 and Berger et al, 1996 for a formal description of these algorithms.

[10] In other words, given known data statistics, construct a model that best represents them but is otherwise as uniform as possible.

[11] Legend: P: Precision; R: Recall; F: F-Measure. NB: naïve Bayes; IIS: Maximum Entropy with Improved Iterative Scaling; GIS: Maximum Entropy with Generalized Iterative Scaling. (Positions/Feature type)

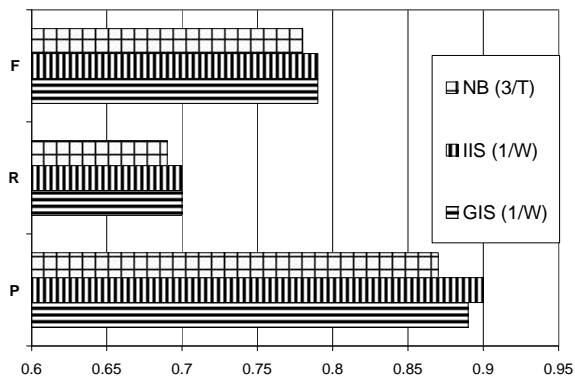

Figure 1. Best metrics for Sociology corpus

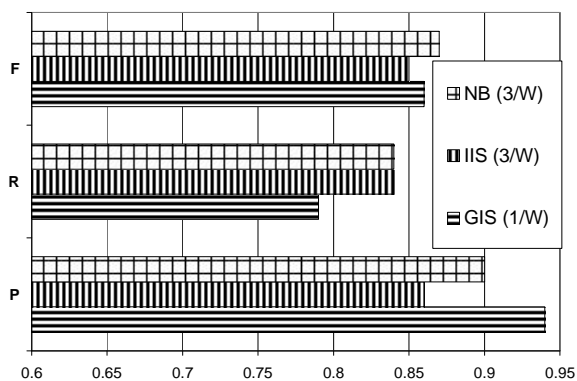

Figure 2. Best metrics for Histology corpus

Although our tests using collocations showed that structural regularities would perform well, our intuitions about improvement using more features (more positions to the right or left of the lexical markers) or a more grammatically restricted environment (surrounding POS tags), turned out to be overly optimistic. Nevertheless, stochastic approaches that used short-range features did perform in line with the hand-coded approach. Both Knowledge-Engineering and supervised learning approaches were adequate for initial filtering of metalinguistic sentences, although learning algorithms might allow easier transport of systems into new domains.

## 5 From EMOs to metalinguistic databases

After EMOs were obtained, POS tagging, shallow parsing and limited PP-attachment are performed. Resulting chunks were tagged as *Autonyms, Agents, Markers, Anaphoric elements* or *Noun Chunks*, using heuristics based on syntactic, pragmatic and argument structure of lexica in the extraction patterns, as well as on FrameNet data in *Name conferral* and *Name*

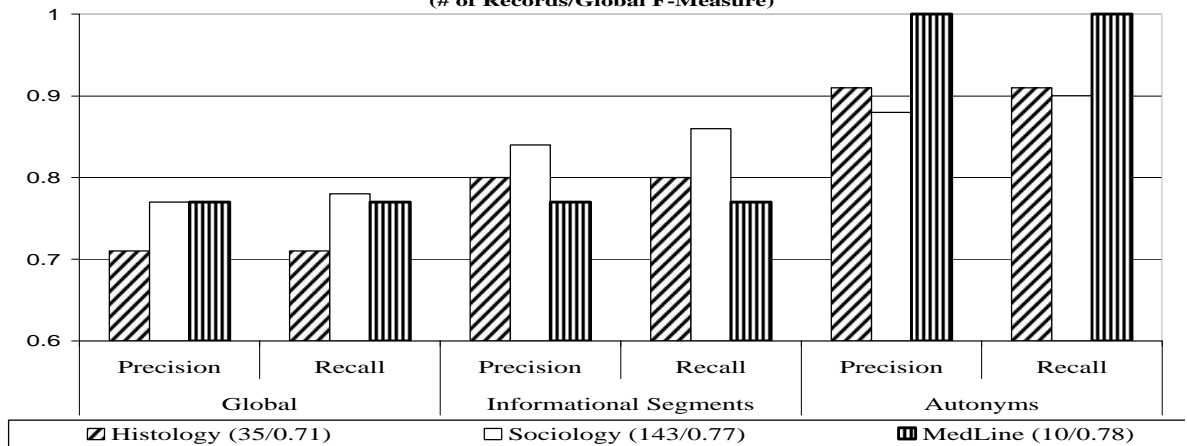

Figure 4. Metrics for 3 corpora
(# of Records/Global F-Measure)

*bearing* frames. Next, a predicate processing phase selected the most likely surface realization for informational segments, autonyms and makers-operators, and proceeded to fill out the templates of the database. As mentioned earlier, informational segments present many realizations far from the completeness and conciseness of lexicographic entries. In fact, they may show up as full-fledged clauses (6), as inter- or intra-sentential anaphoric elements (7 and 8), as sortal information (9), or as an unexpressed "existential variable" (logical form $\exists x$) indicating only that certain discourse entity is being introduced (10):

(6) In 1965 the term soliton was coined **to describe waves with this remarkable behaviour**.

(7) This leap brings cultural citizenship in line with **what** has been called the politics of citizenship .

(8) **They** are called "endothermic compounds."

(9) One of the most enduring aspects of all social theories are **those conceptual entities** known as structures or groups.

(10) A **[$x]** so called cell-type-specific TF can be used by closely related cells….

We have not included an anaphora-resolution module in our system, so that examples 7, 8 and 10 only output either unresolved surface elements or variable placeholders.[12] Nevertheless, more common occurrences like example sentence (1) are enough to create MIDs that constitute useful resources for lexicographers. The correct database entry for (1) is presented below.

| Reference | Histology sample # 6 |
| --- | --- |
| Autonym | *tracheae* |
| Information | *fine hollow tubes* |
| Markers/Operators | *known as* |

To better reflect overall performance, we introduced a threshold of similarity of 65% for comparison between a golden standard slot entry and the one obtained by the application.[13] The final processing stage presented metrics shown in Figure 4. Our best numbers for informational segments ranged around 0.85, while the lowest were obtained for the histology corpus, with global precision and recall rates around 0.71, but with high numbers in the autonym identification task (0.91) and midrange ones for the informational segments (0.8). We observed that even though it is assumed that Bio-Medical Sciences have more consolidated vocabularies than Social Sciences, results for the MedLine and histology corpus occupy the extremes in the spectrum, with the sociology one in the middle range. The total number of candidate sentences was not a good predictor of system performance.

The DEFINDER system (Klavans et al., 2001) is to my knowledge the only one fully comparable with MOP, both in scope and goals, but with some significant differences.[14] Taking into account those differences, MOP compares well with the 0.8 precision and 0.75 recall of DEFINDER. While the resulting MOP "definitions" generally do not present high readability or completeness, these informational segments are not meant to be read by laymen, but used by domain lexicographers updating existing glossaries for neological change, or, in machine-readable form, by other applications.

---

[12] For sentence (8) the system might retrieve useful information from a previous one: "*A few have positive enthalpies of formation.*"

[13] Thus, if the autonym or the informational segment is at least 2/3 of the correct response, it is counted as a positive, allowing for expected errors in the PP or acronym attachment algorithms.

[14] DEFINDER examines user-oriented documents

## 6 Discussion and future work

We have chosen to exploit metalinguistic information that is being put forward in text because it can't be assumed to be part of the collective expert-domain competence. In doing so, we have exposed our system to the less predictable lexical environment of leading-edge research literature, the cauldron where knowledge and terminological systems are forged in real time, and where scientific meaning and interpretation are constantly debated, modified and agreed upon. We believe that low recall rates in our tests are in part due to the fact that we are dealing with the wider realm of metalinguistic information, as opposed to structured definitional sentences that have been distilled by an expert for consumer-oriented documents. We have not performed major customization of the system (like enriching the tagging lexicon with medical terms), in order to preserve the ability to use the system across different domains. Domain customization may improve metrics, but at a real cost for portability.

Conventional resources like lexicons and dictionaries compile meaning definitions that are considered stable and widely-shared. They can be seen as repositories of the default, core lexical information for terms used by a research community (that is, the information available to an average, idealized speaker). An MID, on the other hand, might contain the multi-textured real-time data embedded in research papers, and in this sense could be conceptualized as an *anti-dictionary*: a listing of exceptions, special contexts and specific usage of instances where meaning, value or pragmatic conditions have been spotlighted by discourse for cognitive reasons. Applications that rely on lookup on previously compiled resources would miss some of the data from EMOs, where the term is put forward for the first time, or where important, context-sensitive information about the terms is provided. MIDs cannot be viewed as end-user products, but as semi-structured resources (midway between raw corpora and structured lexical bases) that have to be further processed to convert them into usable data sources. We might better characterize them as auxiliary lexical knowledge resources, more than core lexical references. Lexicographers and terminologist can use them as tools for their own labour-intensive work of reviewing and compiling special-domain vocabularies.

MIDs could, in principle, also supply new interpretation rules in AI applications when inferences won't succeed because the state of the lexico-conceptual system has changed.[15] A neologism or a word in an unexpected technical sense could stump a NLP system that assumes it will be able to use the default information from a machine-readable dictionary or TKB. The kind of sortal information implicit in many definitions can also help improve anaphora resolution, semantic typing, acronym identification or bootstrapping of ontologies and taxonomies (Hearst, 1992; Condamines & Rebeyrolle, 2001; Pustejovsky et al., 2002; Malaisé et al. 2004). Although our approach might miss some of the important conceptual relations between terms, many of the MIDs we have obtained using language-centred contexts are rich sources of information. In addition, terminological information can be more specific than that obtainable by glossary lookup, and might be better suited for the interpretation of certain texts. The locality of such information can be seen as advantageous for specialized lexicography. Another area where non-standard information could prove useful is the study of the evolution of scientific sublanguages and the knowledge embodied by them. Changes in the

---

with fully-developed definitions for the layman. MOP focuses on leading-edge research papers that present less predictable templates. DEFINDER's qualitative evaluation criteria includes readability, usefulness and completeness, as judged by lay subjects, criteria which we have not adopted here, nor have we determined coverage against existing on-line dictionaries.

[15] When interpreting text, regular lexical information is applied by default under normal conditions, but more specific pragmatic or discursive information can override it if necessary, or if context demands so (Lascarides & Copestake, 1995).

conceptual and terminological configuration of a discipline might be traced and be better understood by the dynamical updates reflected in these databases. The next table shows a small sample, taken from our corpora, of that information's potential range, for some key concepts in the sociology domain.

| Terms | Informational segments from EMOs |
|---|---|
| Family | extend the meaning to include same-sex couples, single-parents, nannies, adoptive and step children, and so on |
| Family | two adults of opposite sex, married to each other, and living with their common children |
| Identity | There are two typical contexts |
| Identity | an emotional attachment and a sense of belonging of a semi-sacred kind |
| Nationalism | used here, deliberately, to describe both aspects of the phenomenon |
| Nationalism | is used for both of these things - world view and activism |

## 6.1 Conclusions

The implementation we have described here undoubtedly shows room for improvement: adding more patterns for better overall recall rates, deeper parsing for more accurate semantic typing of sentence arguments, etc. Also, the question of which learning algorithms can better perform the initial filtering of EMO candidates is still very much an open issue. We believe that the real challenge facing work such as this one lies not in retrieving EMOs from text to populate a MID, but in the successful formalization of heterogeneous linguistic information into a robust and manageable data structure. An effective and efficient computational representation of such diverse information is not trivial.

Nevertheless, we believe that applications focused on metalanguage, like the MOP system described here, can be very helpful for Terminology and lexicography, and that a MID's role would not be to replace, but to enrich and complement, Terminological Knowledge Bases.